\def\year{2018}\relax
\documentclass[letterpaper]{article} 
\usepackage{aaai18}  
\usepackage{url}  
\usepackage{times}
\usepackage[multiple]{footmisc}
\usepackage{helvet}
\usepackage{courier}
\usepackage{graphicx}
\usepackage{amsmath}
\usepackage{amssymb}
\usepackage{algorithm}
\usepackage{multirow} 
\usepackage{algorithmic}
\usepackage{booktabs,caption,fixltx2e}
\usepackage[flushleft]{threeparttable}
\usepackage{subfig}
\usepackage{amsthm}
\usepackage{threeparttable}
\usepackage{placeins}

\newcommand\Mark[1]{\textsuperscript#1}

\DeclareMathOperator*{\argmax}{\arg\!\max}
\title{Unified Spectral Clustering with Optimal Graph}
\author{Zhao Kang\Mark{1}, Chong Peng\Mark{2}, Qiang Cheng\Mark{3} and Zenglin Xu\Mark{1}\thanks{Corresponding author.}\\
\Mark{1}School of Computer Science and Engineering, 
University of Electronic Science and Technology of China\\
\Mark{2}Department of Computer Science, Southern Illinois University, Carbondale, USA  \\
\Mark{3}Institute of Biomedical Informatics and Department of Computer Science, University of Kentucky USA  \\
zkang@uestc.edu.cn, pchong@siu.edu, qiang.cheng@uky.edu, zlxu@uestc.edu.cn}
\nocopyright
\begin{document}

\maketitle

\begin{abstract}
Spectral clustering has found extensive use in many areas. Most traditional spectral clustering algorithms work in three separate steps: similarity graph construction; continuous labels learning; discretizing the learned labels by k-means clustering. Such common practice has two potential flaws, which may lead to severe information loss and performance degradation. First, predefined similarity graph might not be optimal for subsequent clustering. It is well-accepted that similarity graph highly affects the clustering results.  To this end, we propose to automatically learn similarity information from data and simultaneously consider the constraint that the similarity matrix has exact c connected components if there are c clusters. Second, the discrete solution may deviate from the spectral solution since k-means method is well-known as sensitive to the initialization of cluster centers. In this work, we transform the candidate solution into a new one that better approximates the discrete one. Finally, those three subtasks are integrated into a unified framework, with each subtask iteratively boosted by using the results of the others towards an overall optimal solution. It is known that the performance of a kernel method is largely determined by the choice of kernels. To tackle this practical problem of how to select the most suitable kernel for a particular data set, we further extend our model to incorporate multiple kernel learning ability. Extensive experiments demonstrate the superiority of our proposed method as compared to existing clustering approaches.
\end{abstract}
\section{Introduction}
Clustering is a fundamental technique in machine learning, pattern recognition, and data mining \cite{huang2017robust}. In past decades, a variety of clustering algorithms have been developed, such as k-means clustering and spectral clustering.    

With the benefits of simplicity and effectiveness, k-means clustering algorithm is often adopted in various real-world problems. To deal with the nonlinear structure of many practical data sets, kernel k-means (KKM) algorithm has been developed \cite{scholkopf1998nonlinear}, where data points are mapped through a nonlinear transformation into a higher dimensional feature space in which the data points are linearly separable. KKM usually achieves better performance than the standard k-means. To cope with noise and outliers, robust kernel k-means (RKKM) \cite{du2015robust} algorithm has been proposed. In this approach, the squared $\ell_2$ norm of error construction term is replaced by $\ell_{2,1}$ norm. RKKM demonstrates superior performance on a number of benchmark data sets. The performance of such model-based methods heavily depends on whether the data fit the model. Unfortunately, in most cases, we do not know the distribution of data in advance. To some extent, this problem is alleviated by multiple kernel learning. Moreover, there is no theoretical result on how to choose the similarity graph \cite{von2007tutorial}.

Spectral clustering is another widely used clustering method \cite{kumar2011co}. It enjoys the advantage of exploring the intrinsic data structures by exploiting the different similarity graphs of data points \cite{yang2015multitask}. 
There are three kinds of similarity graph constructing strategies: k-nearest-neighborhood (knn); $\epsilon$-nearest-neighborhood; The fully connected graph. Here, some open issues arise \cite{huang2015new}: 1) how to choose a proper neighbor number $k$ or radius $\epsilon$; 2) how to select an appropriate similarity metric to measure the similarity among data points; 3) how to counteract the adverse effect of noise and outliers; 4) how to tackle data with structures at different scales of size and density. Unfortunately, all of these issues heavily influence the clustering results \cite{zelnik2004self}.  Nowadays, many data are often high dimensional, heterogeneous, and without prior knowledge, and it is therefore a fundamental challenge to define a pairwise similarity graph for effective spectral clustering.

Recently, \cite{zhu2014constructing} construct robust affinity graphs for spectral clustering by identifying discriminative features. It adopts a random forest approach based on the motivation that tree leaf nodes contain discriminative data partitions, which can be exploited to capture subtle and weak data affinity. This approach shows better performance than other state-of-the-art methods including the Euclidean-distance-based knn \cite{wang2008active}, dominant neighbourhoods \cite{pavan2007dominant}, consensus of knn \cite{premachandran2013consensus}, and non-metric based unsupervised manifold forests \cite{pei2013unsupervised}. 

The second step of spectral clustering is to use the spectrum of the similarity graph to reveal the cluster structure of the data. Due to the discrete constraint on the cluster labels, this problem is NP-hard. To obtain a feasible approximation solution, spectral clustering solves a relaxed version of this problem, i.e., the discrete constraint is relaxed to allow continuous values. It first performs eigenvalue decomposition on the Laplacian matrix to generate an approximate indicator matrix with continuous values. Then, k-means is often implemented to produce final clustering labels \cite{huang2013spectral}. Although this approach has been widely used in practice, it may exhibit poor performance since the k-means method is well-known as sensitive to the initialization of cluster centers \cite{ng2002spectral}. 

To address the aforementioned problems, in this paper, we propose a unified spectral clustering framework. It jointly learns the similarity graph from the data and the discrete clustering labels by solving an optimization problem, in which the continuous clustering labels just serve as intermediate products. To the best of our knowledge, this is the first work that combine the three steps into a single optimization problem. As we show later, it is not trivial to unify them. 
The contributions of our work are as follows:
\begin{enumerate}
\item{Rather than using predefined similarity metrics, the similarity graph is adaptively learned from the data in a kernel space. By combining similarity learning with subsequentl clustering into a unified framework, we can ensure the optimality of the learned similarity graph. }
\item{Unlike existing spectral clustering methods that work in three separate steps, we simultaneously learn similarity graph, continuous labels, and discrete cluster labels. By leveraging the inherent interactions between these three subtasks, they can be boosted by each other.}
\item{Based on our single kernel model, we further extend it to have the ability to learn the optimal combination of multiple kernels. }
\end{enumerate}

\textbf{Notations.} Given a data set $[x_1,x_2,\cdots,x_n]$, we denote $X\in\mathcal{R}^{m\times n}$ with $m$ features and $n$ samples. Then the $i$-th sample and $(i,j)$-th element of matrix $X$ are denoted by $x_i\in \mathcal{R}^{m\times 1}$ and $x_{ij}$, respectively. The $\ell_2$-norm of a vector \textbf{$x$} is defined as \textbf{$\|x\|^2=x^\top\cdot x$}, where $\top$ means transpose. The squared Frobenius norm is denoted by $\|X\|_F^2=\sum_{ij}x_{ij}^2$. The $\ell_1$-norm of matrix $X$ is defined as the absolute summation of its entries, i.e., $\|X\|_1=\sum_i\sum_j |x_{ij}|$. $I$ denotes the identity matrix. Tr($\cdot$)  is the trace operator. $ Z\ge 0$ means all the elements of $Z$ are nonnegative.

\section{Preliminary Knowledge}
\label{related}
\subsection{Sparse Representation}
Recently, sparse representation, which assumes that each data point can be reconstructed as a linear combination of the other data points, has shown its power in many tasks \cite{cheng2010learning,peng2016feature}. It often solves the following problem:
 \begin{equation}
\label{sparse}
\min_Z \|X- XZ\|_F^2+\alpha\|Z\|_1,\hspace{.1cm} 
s.t. \hspace{.1cm}  Z\ge0, \hspace{.1cm} diag(Z)=0,
\end{equation}
where $ \alpha>0$ is a balancing parameter. 
Eq. (\ref{sparse}) simultaneously determines both the neighboring samples of a data point and the corresponding weights by the sparse reconstruction from the remaining samples. In principle, more similar points should receive bigger weights and the weights should be smaller for less similar points. Thus $Z$ is also called similarity graph matrix \cite{kang2015robust}. In addition, sparse representation enjoys some nice properties, e.g., the robustness to noise and datum-adaptive ability \cite{huang2015new}. On the other hand, model (\ref{sparse}) has a drawback, i.e., it does not consider nonlinear data sets where data points reside in a union of manifolds \cite{kang2017kernel}.

\subsection{Spectral Clustering}
Spectral clustering requires Laplacian matrix $L\in \mathcal{R}^{n\times n}$ as an input, which  is computed as $L=D-\frac{Z^\top+Z}{2}$, where $D\in \mathcal{R}^{n\times n}$ is a diagonal matrix with the $i$-th diagonal element $\sum_j \frac{z_{ij}+z_{ij}}{2}$. In traditional spectral clustering methods, similarity graph $Z\in \mathcal{R}^{n\times n}$ is often constructed in one of the three ways aforementioned.  Supposing there are $c$ clusters in the data $X$, spectral clustering solves the following problem:
\begin{equation}
\label{spectral}
\min_{F} Tr(F^\top LF),\quad s.t. \quad F\in Idx,
\end{equation}
where $F=[f_1,f_2,\cdots,f_n]^\top\in \mathcal{R}^{n\times c}$ is the cluster indicator matrix and $F\in Idx$ represents the clustering label vector of each point $f_i\in\{0,1\}^{c\times 1}$ contains one and only one element ``1'' to indicate the group membership of $x_i$. Due to the discrete constraint on $F$, problem (\ref{spectral}) is NP-hard. In practice, $F$ is relaxed to allow continuous values and solve 
\begin{equation}
\label{spectral2}
\min_{P} Tr(P^\top LP),\quad s.t. \quad P^\top P=I,
\end{equation}
where $P\in \mathcal{R}^{n\times c}$ is the relaxed continuous clustering label matrix, and the orthogonal constraint is adopted to avoid trivial solutions. The optimal solution is obtained from the $c$ eigenvectors of $L$ corresponding to the $c$ smallest eigenvalues. After obtaining $F$, traditional clustering method, e.g., k-means, is implemented to obtain discrete cluster labels \cite{huang2013spectral}. 

Although this three-steps approach provides a feasible solution, it comes with two potential risks. First, since the similarity graph computation is independent of the subsequent steps, it may be far from optimal. As we discussed before, the clustering performance is largely determined by the similarity graph. Thus, final results may be degraded. Second, the final solution may unpredictably deviate from the ground-truth discrete labels \cite{yang2016unified}. To address these problems, we propose a unified spectral clustering model.
\section{Spectral Clustering with Single Kernel} 
\label{single}
\subsection{Model}
One drawback of Eq. (\ref{sparse}) is that it assumes that all the points lie in a union of independent or disjoint subspaces and are noiseless. In the presence of dependent subspaces, nonlinear manifolds and/or data errors, it may select points from different structures to represent a data point and makes the representation less informative \cite{elhamifar2009sparse}. It is recognized that nonlinear data may represent linearity when mapped to an implicit, higher-dimensional space via a kernel function. To fully exploit data information, we formulate Eq. (\ref{sparse}) in a general manner with a kernelization framework.

Let $\phi: \mathcal{R}^D\rightarrow\mathcal{H}$ be a kernel mapping the data samples from the input space to a reproducing kernel Hilbert space $\mathcal{R}$. Then $X$ is transformed to $\phi(X)=[\phi(x_1),\cdots,\phi(x_n)]$. The kernel similarity between data samples $x_i$ and $x_j$ is defined through a predefined kernel as $K_{x_i,x_j}=<\phi(x_i),\phi(x_j)>$. By applying this kernel trick, we do not need to know the transformation $\phi$. In the new space, Eq. (\ref{sparse}) becomes
\cite{zhang2010general}
\begin{equation}
\begin{split}
&\min_{Z} \|\phi(X)-\phi(X)Z\|_F^2+\alpha \|Z\|_1,\\
\Longleftrightarrow&\min_{Z} Tr(\phi(X)^T\phi(X)-\phi(X)^T\phi(X) Z\\
&-Z^T\phi(X)^T\phi(X)+Z^T\phi(X)^T\phi(X) Z)+\alpha \|Z\|_1,\\
\Longleftrightarrow
& \min_{Z} Tr(K-2KZ+Z^\top KZ)+\alpha \|Z\|_1,\\
&s.t. \quad Z\ge0, \quad diag(Z)=0,
\label{kernelsparse}
\end{split}
\end{equation}
This model recovers the linear relations among the data in the new space, and thus the nonlinear relations in the original representation. Eq. (\ref{kernelsparse}) is more general than Eq. (\ref{sparse}) and is supposed to learn arbitrarily shaped data structure. Moreover, Eq. (\ref{kernelsparse}) goes back to Eq. (\ref{sparse}) when a linear kernel is applied.

To fulfill the clustering task, we propose our spectral clustering with single kernel (SCSK) model as following:
\begin{equation}
\begin{split}
&  \min_{Z,F,P,Q} \underbrace{Tr(K-2KZ+Z^\top KZ)+\alpha \|Z\|_1}_\text{similarity learning}\\
&+\underbrace{\beta Tr(P^\top LP)}_\text{continuous label learning}+\underbrace{\gamma \|F-PQ\|_F^2}_\text{discrete label learning},\\
&s.t. \quad Z\ge0, \quad diag(Z)=0,\\
& \quad P^\top P=I,\quad Q^\top Q=I,\quad F\in Idx,
\label{scsk}
\end{split}
\end{equation}
where $\alpha$, $\beta$, and $\gamma$ are penalty parameters, and $Q$ is a rotation matrix. Due to the spectral solution invariance property \cite{yu2003multiclass}, for any solution $P$, $PQ$ is another solution. The purpose of the last term is to find a proper orthonormal $Q$ such that the resulting $PQ$ is close to the real discrete clustering labels. In Eq. (\ref{scsk}), the similarity graph and the final discrete clustering labels are automatically learned from the data. Ideally, whenever data points $i$ and $j$ belong to different clusters, we must have $z_{ij}=0$ and it is also true vice versa. That is to say, we have $z_{ij}\neq 0$ if and only if data points $i$ and $j$ are in the same cluster, or, equivalently $f_i=f_j$. Therefore, our unified framework Eq. (\ref{scsk}) can exploit the correlation between the similarity matrix and the labels. Because of the feedback of inferred labels to induce the similarity matrix and vice versa, we say that our clustering framework has a self-taught property. 
%

In fact, Eq. (\ref{scsk}) is not a simple unification of the pipeline of steps. It learns a similarity graph with optimal structure for clustering. Ideally, $Z$ should have exactly $ c$ connected components if there are $c$ clusters in the data set \cite{kang2017twin}. This is to say that the Laplacian matrix $L$ has $c$ zero eigenvalues \cite{mohar1991laplacian}, i.e., the summation of the smallest $c$ eigenvalues is zero. To ensure the optimality of the similarity graph, we can minimize $\sum_{i=1}^{c}\sigma_i(L)$. According to Ky Fan's theorem \cite{fan1949theorem}, $\sum_{i=1}^{c}\sigma_i(L)=\min\limits_{P^\top P=I}Tr(P^\top LP)$. Therefore, the spectral clustering term, i.e., the second term in Eq. (\ref{scsk}), will ensure learned $Z$ is optimal for clustering. 
\subsection{Optimization}
To efficiently and effectively solve Eq. (\ref{scsk}), we design an alternated iterative method.

\noindent
\textbf{Computation of Z:} With $F$, $P$, $Q$ fixed, the problem is reduced to 
\begin{equation}
\begin{split}
& \min_{Z} Tr(K-2KZ+Z^\top KZ)+\alpha \|Z\|_1+\beta Tr(P^\top LP),\\
&s.t. \quad Z\ge0, \quad diag(Z)=0.
\end{split}
\end{equation}
We introduce an auxiliary variable $S$ to make above objective function separable and solve the following equivalent problem:
\begin{equation}
\begin{split}
& \min_{Z} Tr(K-2KZ+Z^\top KZ)+\alpha \|S\|_1+\beta Tr(P^\top LP),\\
&s.t. \quad Z\ge0, \quad diag(Z)=0, \quad S=Z.
\end{split}
\end{equation}
This can be solved by using the augmented Lagrange multiplier (ALM) type of method. We turn to minimizing the following augmented Lagrangian function:
\begin{equation}
\begin{split}
\mathcal{L}(S,Z,Y)=&Tr(K-2KZ+Z^\top KZ)+\alpha \|S\|_1\\
&+\beta Tr(P^\top LP)+\frac{\mu}{2}\|S-Z+\frac{Y}{\mu}\|_F^2,
\label{lag}
\end{split}
\end{equation}
where $\mu>0$ is the penalty parameter and $Y$ is the Lagrange multiplier. This problem can be minimized with respect to $S$, $Z$, and $Y$ alternatively, by fixing the other variables. 

For $S$, by letting $H=Z-\frac{Y}{\mu}$, it can be updated element-wisely as below： 
\begin{equation}
S_{ij}=max(|H_{ij}|-\alpha/\mu,0)\cdot sign(H_{ij}).
\label{updatew}
\end{equation}
For $Z$, by letting $E=S+\frac{Y}{\mu}$, it can be updated column-wisely as:
\begin{equation}
\min_{Z_i} Z_i^T(\frac{\mu}{2} I +K)Z_i+(\frac{\beta}{2}d_i^T-\mu E_i^T-2K_{i,:}) Z_i,
\label{updatez}
\end{equation}
where $d_i\in \mathcal{R}^{n\times 1}$ is a vector with the $j$-th element $d_{ij}$ being $d_{ij}=\|P_{i,:}-P_{j,:}\|^2$. It is easy to obtain $Z_i$ by setting the derivative of Eq. (\ref{updatez}) w.r.t. $Z_i$ to be zero. 

\noindent
\textbf{Computation of P:} With $F$, $Z$, and $Q$ fixed, it is equivalent to solving
\begin{equation}
\min_P \beta Tr(P^\top LP)+\gamma\|F-PQ\|_F^2\hspace{.1cm}s.t.\hspace{.1cm} P^\top P=I.
\label{updatep}
\end{equation}
The above problem with orthogonal constraint can be efficiently solved by the algorithm proposed by Wen and Yin \cite{wen2013feasible}.\\
\noindent
\textbf{Computation of Q:} With $F$, $Z$, and $P$ fixed, we have
\begin{equation}
\min_Q \|F-PQ\|_F^2\quad s.t.\quad Q^\top Q=I.
\end{equation}
It is the orthogonal Procrustes problem \cite{schonemann1966generalized}, which admits a closed-form solution. The solution is
\begin{equation}
Q=UV^\top,
\label{updateq}
\end{equation} 
where $U$ and $V$ are left and right parts of the SVD decomposition of $F^\top P$.\\
\noindent
\textbf{Computation of F:} With $Z$, $P$ and $Q$ fixed, the problem becomes
\begin{equation}
\min_F \|F-PQ\|_F^2,\quad s.t.\quad F\in Idx.
\end{equation}
Note that $Tr(F^\top F)=n$, the above subproblem can be rewritten as below:
\begin{equation}
\max_F Tr(F^\top PQ)\quad s.t.\quad F\in Idx.
\end{equation}
The optimal solution can be easily obtained as follows:
\begin{equation}
   F_{ij}=
\begin{cases}
      1, & \text{$j=\argmax\limits_k\hspace{.1cm} (PQ)_{ik}$}\\
      0, & \text{otherwise}
\end{cases}
\label{updatef}
\end{equation}

The updates of $Z$, $P$, $F$, and $Q$ are coupled with each other, so we could reach an overall optimal solution. The details of our SCSK optimization are summarized in Algorithm 1.
\begin{algorithm}[!tb]
\small
\caption{The algorithm of SCSK }
\label{alg1}
 {\bfseries Input:} Kernel matrix $K$, parameters  $\alpha>0$, $\beta>0$, $\gamma>0$, $\mu>0$.\\
{\bfseries Initialize:} Random matrices $Z$, $P$, and $Q$. $Y=0$ and $F=0$.\\
 {\bfseries REPEAT}
\begin{algorithmic}[1]
 \STATE Update $S$ according to Eq. (\ref{updatew}).
\STATE$ S=S-diag(diag(S))$ and $S=max(S, 0)$.
 \STATE Update $Z$ according to Eq. (\ref{updatez}).
\STATE $Y=Y+\mu(S-Z)$.
\STATE Update $P$ by solving the problem of Eq. (\ref{updatep}).
   \STATE Update $Q$ according to Eq. (\ref{updateq}).
\STATE Update $F$ according to Eq. (\ref{updatef}). 
\end{algorithmic}
\textbf{ UNTIL} {stopping criterion is met.}
\end{algorithm}

\subsection{Complexity Analysis}
With our optimization strategy, the updating of $S$ requires $\mathcal{O}(n^2)$ complexity. The quadratic program can be solved in polynomial time. The solution of $Q$ involves SVD and its complexity is $\mathcal{O}(nc^2+c^3)$. To update $P$, we need $\mathcal{O}(nc^2+c^3)$. The complexity for $F$ is $\mathcal{O}(nc^2)$. Note that the number of clusters $c$ is often a small number. Therefore, the main computation load is from solving $Z$, which involves matrix inversion. Fortunately, $Z$ is solved in parallel.

\section{Spectral Clustering with Multiple Kernels}
\label{multiple}
\subsection{Model}
Although the model in Eq. (\ref{scsk}) can automatically learn the similarity graph matrix  and discrete cluster labels, its performance will strongly depend on the choice of kernels. It is often impractical to exhaustively search for the most suitable kernel. Moreover, real world data sets are often generated from different sources along with heterogeneous features. Single kernel method may not be able to fully utilize such information. Multiple kernel learning has the ability to integrate complementary information and identify a suitable kernel for a given task. Here we present a way to learn an appropriate consensus kernel from a convex combination of a number of predefined kernel functions.

Suppose there are a total number of $r$ different kernel functions $\{K^i\}_{i=1}^{r}$. An augmented Hilbert space can be constructed by  using the mapping of $\tilde{\phi}(x)=[\sqrt{w_1}\phi_1(x),$ $\sqrt{w_2}\phi_2(x),...,\sqrt{w_r}\phi_r(x)]^\top$ with different weights $\sqrt{w_i}(w_i\ge 0)$. Then the combined kernel $K_w$ can be represented as \cite{zeng2011feature}
\begin{equation}
\label{ukernel}
K_w(x,y)=<\phi_w(x),\phi_w(y)>=\sum\limits_{i=1}^r w_iK^i(x,y).
\end{equation} 
Note that the convex combination of the positive semi-definite kernel matrices  $\{K^i\}_{i=1}^{r}$ is still a positive semi-definite kernel matrix. Thus the combined kernel still satisfies Mercer's condition. Then our proposed method of spectral clustering with multiple kernels (SCMK) can be formulated as
\begin{equation}
\begin{split}
  \min_{Z,F,P,Q,w} &Tr(K_w-2K_wZ+Z^\top K_wZ)+\alpha \|Z\|_1+\\
&\beta Tr(P^\top LP)+\gamma \|F-PQ\|_F^2,\\
&s.t. \quad Z\ge0, \quad diag(Z)=0,\\
& \quad P^\top P=I,\quad Q^\top Q=I,\quad F\in Idx,\\
&\quad K_w=\sum\limits_{i=1}^r w_iK^i, \quad\sum\limits_{i=1}^r \sqrt{w_i}=1,\quad w_i\ge 0.
\label{scmk}
\end{split}
\end{equation}
Now above model will learn the similarity graph, discrete clustering labels, and kernel weights by itself. By iteratively updating $Z$, $F$, and $w$, each of them will be iteratively refined according to the results of the others. 
\subsection{Optimization }
In this part, we show an efficient and effective algorithm to iteratively and alternatively solve Eq. (\ref{scmk}).

\textbf{$w$ is fixed:} Update other variables when $w$ is fixed: We can directly calculate $K_w$, and the optimization problem is exactly Eq. (\ref{scsk}). Thus we just need to use Algorithm 1 with $K_w$ as the input kernel matrix.

\textbf{Update $w$:} Optimize with respect to $w$ when other variables are fixed: Solving Eq. (\ref{scmk}) with respect to $w$ can be rewritten as \cite{cai2013heterogeneous}
\begin{equation}
\begin{split}
\label{optie}
&\min_w \sum\limits_{i=1}^r w_i h_i  \\
& s.t.\quad  \sum\limits_{i=1}^r \sqrt{w_i}=1, \quad w_i\ge 0, 
\end{split}
\end{equation}
where 
\begin{equation}
\label{h}
h_i=Tr(K^i-2K^iZ+Z^\top K^iZ).
\end{equation}
The Lagrange function of Eq. (\ref{optie}) is 
\begin{equation}
\mathcal{J}(w)=w^\top h+\gamma (1-\sum_{i=1}^r\sqrt{w_i}).
\end{equation}
By utilizing the Karush-Kuhn-Tucker (KKT) condition with $\frac{\partial \mathcal{J}(w)}{\partial w_i}=0$ and the constraint $\sum\limits_{i=1}^r \sqrt{w_i}=1$, we obtain the solution of $w$ as follows:
\begin{equation}
\label{weight}
w_i=\left(h_i \sum_{j=1}^r \frac{1}{h_j}\right)^{-2}.
\end{equation}
We can see that $w$ is closely related to $Z$. Therefore, we could obtain both optimal similarity matrix $Z$ and kernel weight $w$. We summarize the optimization process of Eq. (\ref{scmk}) in Algorithm 2.
\begin{algorithm}[!tb]
\caption{The algorithm of SCMK }
\label{alg2}
 {\bfseries Input:} A set of kernel matrix $\{K^i\}_{i=1}^r$, parameters  $\alpha>0$, $\beta>0$, $\gamma>0$, $\mu>0$.\\
{\bfseries Initialize:} Random matrices $Z$, $P$, and $Q$. $Y=0$ and $F=0$. $w_i=1/r$.\\
 {\bfseries REPEAT}
\begin{algorithmic}[1]
\STATE Calculate $K_w$ by Eq. (\ref{ukernel}).
\STATE Do steps 1-7 in Algorithm 1.
\STATE Calculate $h$ by Eq. (\ref{h}).
\STATE Calculate $w$ by Eq. (\ref{weight}).
\end{algorithmic}
\textbf{ UNTIL} {stopping criterion is met.}
\end{algorithm}

%
%


\section{Experiments}
\label{exp}
\begin{table}[!htb]
\centering
\caption{Description of the data sets}
\label{data}
\renewcommand{\arraystretch}{1.1}
\begin{tabular}{|l|c|c|c|}
\hline
&\textrm{\# instances}&\textrm{\# features}&\textrm{\# classes}\\\hline
\textrm{YALE}&165&1024&15\\\hline
\textrm{JAFFE}&213&676&10\\\hline
\textrm{ORL}&400&1024&40\\\hline
\textrm{AR}&840&768&120\\\hline
\textrm{COIL20}&1440&1024&20\\\hline
\textrm{BA}&1404&320&36\\\hline
\textrm{TR11}&414&6429&9\\\hline
\textrm{TR41}&878&7454&10\\\hline
\textrm{TR45}&690&8261&10\\\hline
\textrm{TDT2}&9394&36771&30\\\hline
\end{tabular}
\end{table}

\subsection{Data Sets}
There are altogether ten real benchmark data sets used in our experiments. Table \ref{data} summarizes the statistics of these data sets. Among them, the first six are image data, and the other four are text corpora\footnote{http://www-users.cs.umn.edu/~han/data/tmdata.tar.gz}\footnote{http://www.cad.zju.edu.cn/home/dengcai/Data/TextData.html}.

The six image data sets consist of four famous face databases (ORL\footnote{http://www.cl.cam.ac.uk/research/dtg/attarchive/facedatabase.html}, YALE\footnote{http://vision.ucsd.edu/content/yale-face-database}, AR\footnote{http://www2.ece.ohio-state.edu/~aleix/ARdatabase.html} and JAFFE\footnote{http://www.kasrl.org/jaffe.html}), a toy image database COIL20\footnote{http://www.cs.columbia.edu/CAVE/software/softlib/coil-20.php}, and a binary alpha digits data set BA\footnote{http://www.cs.nyu.edu/~roweis/data.html}. Specifically, COIL20 contains images of 20 objects. For each object, the images were taken five degrees apart as the object is rotating on a turntable. There are 72 images for each object. Each image is represented by a 1,024-dimensional vector. BA consists of digits of ``0" through ``9" and letters of capital ``A" through ``Z". There are 39 examples for each class. YALE, ORL, AR, and JAFEE contain images of individuals. Each image has different facial expressions or configurations due to times, illumination conditions, and glasses/no glasses.
\subsection{Kernel Design}
To assess the effectiveness of multiple kernel learning, we adopted 12 kernels. They include: seven Gaussian kernels of the form $K(x,y)=exp(-\|x-y\|_2^2/(td_{max}^2))$, where $d_{max}$ is the maximal distance between samples and $t$ varies over the set $\{0.01, 0.0, 0.1, 1, 10, 50, 100\}$; a linear kernel $K(x,y)=x^\top y$; four polynomial kernels $K(x,y)=(a+x^\top y)^b$ with $a\in\{0,1\}$ and $b\in\{2,4\}$. Furthermore, all kernels are rescaled to $[0,1]$ by dividing each element by the largest pair-wise squared distance.

\captionsetup{position=top}
\begin{table*}[!ht]
\centering
\tiny
\renewcommand{\arraystretch}{1.3}
\setlength{\tabcolsep}{.04pt}
\subfloat[Accuracy(\%)\label{acc}]{
\resizebox{.99\textwidth}{!}{
\begin{tabular}{|l  |c |c| c| c| c|c| c| c| c|c| c| c|c | |c| c| c| c| c}
	\hline
    \tiny{Data}  & \tiny{KKM} & \tiny{KKM-m} & \tiny{SC} & \tiny{SC-m} &\tiny{RKKM} & \tiny{RKKM-m} & \tiny{ClustRF-u} & \tiny{ClustRF-a} &\tiny{SSR} & \tiny{TSEP} & \tiny{TSEP-m}& \tiny{SCSK} & \tiny{SCSK-m} &\tiny{MKKM} & \tiny{AASC} & \tiny{RMKKM}&\tiny{SCMK} \\
     \hline
        \multirow{1}{*}{\tiny{YALE}}  & 47.12&38.97&49.42&40.52&48.09&39.71&57.58&57.58&54.55&62.58&44.60&\textbf{63.05}&52.88&45.70&40.64&52.18&\textbf{63.25}\\
	
		\hline
		\multirow{1}{*}{\tiny{JAFFE}}  & 74.39&67.09&74.88&54.03&75.61&67.98&97.65&98.59&87.32&98.30&73.88&\textbf{99.53}&90.06&74.55&30.35&87.07&\textbf{99.69}\\
		
		\hline
        \multirow{1}{*}{\tiny{ORL}}  & 53.53&45.93&57.96&46.65&54.96&46.88&60.75&62.75&69.00&70.15&41.45&\textbf{74.05}&53.56&47.51&27.20&55.60&\textbf{74.52}\\
		
		\hline
        \multirow{1}{*}{\tiny{AR}}  & 33.02&30.89&28.83&22.22&33.43 &31.20&24.17&35.59&65.00&65.03&46.41&\textbf{78.90}&68.21&28.61&33.23&34.37&\textbf{79.29}\\
				\hline
        \multirow{1}{*}{\tiny{COIL20}}  & 59.49&50.74&67.70&43.65&61.64&51.89&74.44&72.99&76.32&77.68&61.03&\textbf{81.48}&62.59&54.82&34.87&66.65&\textbf{82.21}\\
		\hline
        \multirow{1}{*}{\tiny{BA}}& 41.20&33.66&31.07&26.25&42.17&34.35&39.89&44.01&23.97&45.92&30.75&\textbf{46.02}&31.50&40.52&27.07&43.42&\textbf{45.57}\\
      
       \hline
        \multirow{1}{*}{\tiny{TR11}}  & 51.91&44.65&50.98&43.32&53.03&45.04&29.24&34.54&41.06&71.05&42.08&\textbf{74.22}&55.09&50.13&47.15&57.71&\textbf{74.26}\\
		
		\hline
		\multirow{1}{*}{\tiny{TR41}}  & 55.64&46.34&63.52&44.80&56.76&46.80&53.19&60.93&63.78&69.45&50.17&\textbf{70.17}&53.05&56.10&45.90&62.65&\textbf{70.25}\\
		
		\hline
		\multirow{1}{*}{\tiny{TR45}} & 58.79&45.58&57.39&45.96&58.13&45.69&42.17&48.41&71.45&76.54&51.07&\textbf{77.74}&59.53&58.46&52.64&64.00&\textbf{77.47}\\
		
		\hline
		\multirow{1}{*}{\tiny{TDT2}} &47.05&35.58&52.63&45.26&48.35&36.67&-&-&20.86&54.78&46.35&\textbf{56.04}&45.02&34.36&19.82&37.57&\textbf{56.29}\\
\hline
\end{tabular}}

}\\
\tiny
\renewcommand{\arraystretch}{1.3}
\subfloat[NMI(\%)\label{NMI}]{
\resizebox{.99\textwidth}{!}{
\begin{tabular}{|l  |c |c| c| c| c| c| c|c| c|c| c| c|c | |c| c| c| c| c}
	\hline
    \tiny{Data}  & \tiny{KKM} & \tiny{KKM-m} & \tiny{SC} & \tiny{SC-m} &\tiny{RKKM} & \tiny{RKKM-m} & \tiny{ClustRF-u} & \tiny{ClustRF-a}&\tiny{SSR}& \tiny{TSEP} & \tiny{TSEP-m}& \tiny{SCSK} & \tiny{SCSK-m} &\tiny{MKKM} & \tiny{AASC} & \tiny{RMKKM}&\tiny{SCMK} \\
	\hline
        \multirow{1}{*}{\tiny{YALE}} & 51.34&42.07&52.92&44.79&52.29&42.87&58.76&60.25&57.26&60.13&46.10&\textbf{60.58}&52.72&50.06&46.83&55.58&\textbf{61.04}\\
		
		\hline
		\multirow{1}{*}{\tiny{JAFFE}} & 80.13&71.48&82.08&59.35&83.47&74.01&97.00&98.16&92.93&98.61&71.95&\textbf{99.18}&88.86&79.79&27.22&89.37&\textbf{99.20}\\
		
	\hline
        \multirow{1}{*}{\tiny{ORL}} & 73.43&63.36&75.16&66.74&74.23&63.91&78.69&79.87&84.23&83.28&50.76&\textbf{84.78}&70.93&68.86&43.77&74.83&\textbf{85.21}\\
		
	\hline
        \multirow{1}{*}{\tiny{AR}} & 65.21&60.64&58.37&56.05&65.44 &60.81&57.09&66.64&84.16&84.69&64.63&\textbf{89.61}&80.34&59.17&65.06&65.49&\textbf{89.93}\\
		
		\hline
        \multirow{1}{*}{\tiny{COIL20}}  &74.05&63.57&80.98&54.34&74.63&63.70&83.91&82.26&86.89&84.16&71.36&\textbf{87.03}&72.41&70.64&41.87&77.34&\textbf{86.72}\\
		
	\hline
        \multirow{1}{*}{\tiny{BA}}  &57.25&46.49&50.76&40.09&57.82&46.91&54.66&58.17&30.29&59.47&32.45&\textbf{60.34}&42.91&56.88&42.34&58.47&\textbf{60.55}\\
      
	\hline
        \multirow{1}{*}{\tiny{TR11}}  & 48.88&33.22&43.11&31.39&49.69&33.48&18.97&24.77&27.60&62.71&29.88&\textbf{64.60}&44.48&44.56&39.39&56.08&\textbf{64.89}\\
		\hline
		\multirow{1}{*}{\tiny{TR41}} & 59.88&40.37&61.33&36.60&60.77&40.86&52.63&56.78&59.56&64.07&39.58&\textbf{64.92}&47.97&57.75&43.05&63.47&\textbf{64.89}\\
		
		\hline
		\multirow{1}{*}{\tiny{TR45}}& 57.87&38.69&48.03&33.22&57.86&38.96&38.12&43.70&67.82&70.03&40.17&\textbf{70.75}&50.47&56.17&41.94&62.73&\textbf{70.79}\\
		
	\hline
\multirow{1}{*}{\tiny{TDT2}} &55.28&38.47&52.23&27.16&54.46&42.19&-&-&02.44&57.74&45.38&\textbf{59.25}&48.73&41.36&02.14&47.13&\textbf{58.66}\\
\hline
\end{tabular}}
}\\
\tiny
\renewcommand{\arraystretch}{1.3}
\subfloat[ Purity(\%)\label{purity}]{
\resizebox{.99\textwidth}{!}{
\begin{tabular}{|l  |c |c| c| c| c| c| c|c| c|c| c| c|c | |c| c| c| c| c}
	\hline
    \tiny{Data}  & \tiny{KKM} & \tiny{KKM-m} & \tiny{SC} & \tiny{SC-m} &\tiny{RKKM} & \tiny{RKKM-m} & \tiny{ClustRF-u} & \tiny{ClustRF-a}& \tiny{SSR}& \tiny{TSEP} & \tiny{TSEP-m}& \tiny{SCSK} & \tiny{SCSK-m} &\tiny{MKKM} & \tiny{AASC} & \tiny{RMKKM}&\tiny{SCMK} \\
   	\hline
        \multirow{1}{*}{\tiny{YALE}}  &49.15&41.12&51.61&43.06&49.79&41.74&63.64&63.03&58.18&64.77&55.38&\textbf{65.87}&56.19&47.52&42.33&53.64&\textbf{67.39}\\
	\hline
		\multirow{1}{*}{\tiny{JAFFE}}  & 77.32&70.13&76.83&56.56&79.58&71.82&97.65&98.59&96.24&99.06&77.08&\textbf{99.23}&91.24&76.83&33.08&88.90&\textbf{99.51}\\
	\hline
        \multirow{1}{*}{\tiny{ORL}}  & 58.03&50.42&61.45&51.20&59.60&51.46&67.25&66.00&76.50&76.00&52.39&\textbf{77.02}&57.96&52.85&31.56&60.23&\textbf{78.31}\\
		\hline
        \multirow{1}{*}{\tiny{AR}}  & 35.52&33.64&33.24&25.99&35.87 &33.88&40.71&46.79&69.52&72.44&57.25&\textbf{83.08}&70.69&30.46&34.98&36.78&\textbf{83.20} \\
		\hline
        \multirow{1}{*}{\tiny{COIL20}}  & 64.61&55.30&69.92&46.83&66.35&56.34&80.83&77.71&\textbf{89.03}&84.03&74.89&84.24&75.58&58.95&39.14&69.95&\textbf{83.78}\\
	\hline
        \multirow{1}{*}{\tiny{BA}} & 44.20&36.06&34.50&29.07&45.28&36.86&41.95&49.14&40.85&55.03&43.07&\textbf{55.49}&40.45&43.47&30.29&46.27&\textbf{55.72}\\
	\hline
        \multirow{1}{*}{\tiny{TR11}}  & 67.57&56.32&58.79&50.23&67.93&56.40&35.75&49.76&85.02&85.95&63.15&\textbf{86.25}&63.36&65.48&54.67&72.93&\textbf{85.84}\\
	\hline
		\multirow{1}{*}{\tiny{TR41}} & 74.46&60.00&73.68&56.45&74.99&60.21&55.58&65.60&75.40&77.02&56.33&\textbf{78.53}&57.19&72.83&62.05&77.57&\textbf{78.49}\\
		\hline
		\multirow{1}{*}{\tiny{TR45}} & 68.49&53.64&61.25&50.02&68.18&53.75&45.51&57.83&\textbf{83.62}&77.28&60.52&79.70&61.06&69.14&57.49&75.20&\textbf{79.78}\\
	\hline
\multirow{1}{*}{\tiny{TDT2}} &52.79&49.26&50.39&42.81&62.13&52.60&-&-&46.79&67.75&56.07&\textbf{70.69}&64.53&54.89&21.73&60.02&\textbf{72.84}\\
\hline
\end{tabular}
}}
\caption{Clustering results obtained on benchmark data sets. '-m' denotes the average performance on the 12 kernels. Both the best results for single kernel and multiple kernel methods are highlighted in boldface. \label{clusterres}}
\end{table*}


\subsection{Comparison Algorithms}
\captionsetup{position=bottom}
\begin{figure*}[!htbp]
\centering
\subfloat[$\gamma=10^{-5}$\label{$gamma=1e-5$}]{\includegraphics[width=.33\textwidth]{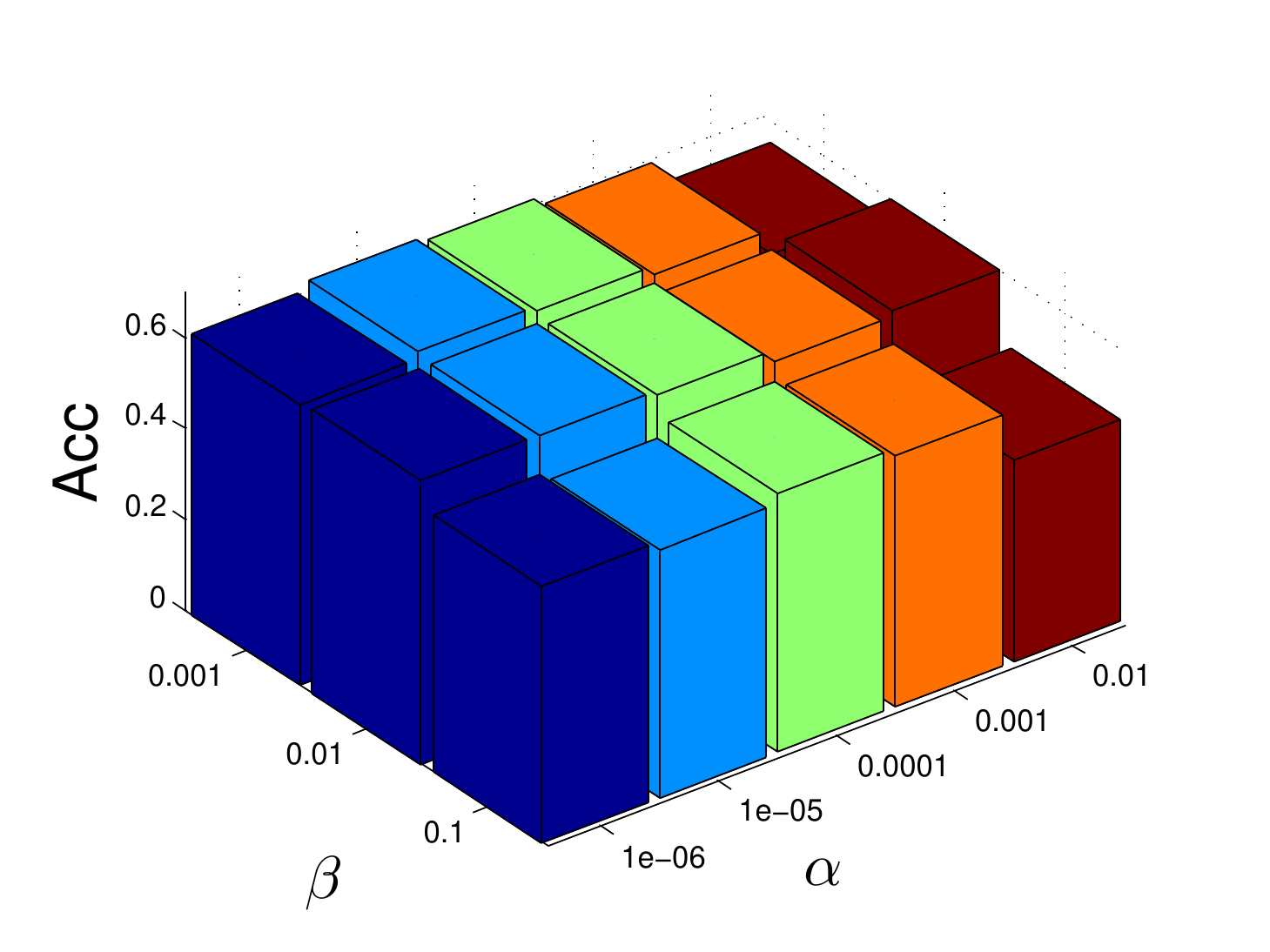}}
\subfloat[$\gamma=10^{-4}$\label{$gamma=1e-4$}]{\includegraphics[width=.33\textwidth]{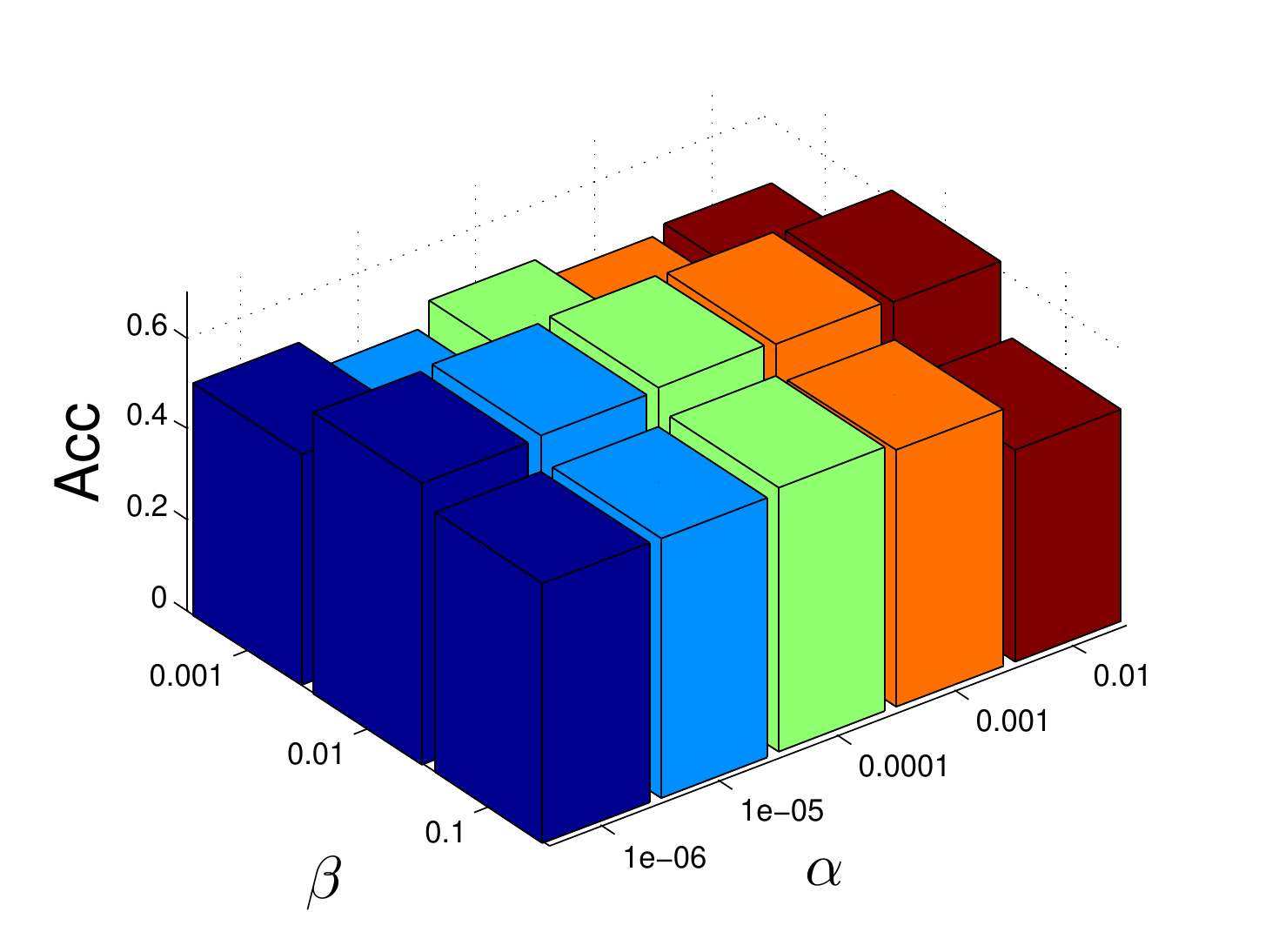}}
\subfloat[$\gamma=10^{-3}$\label{$gamma=1e-3$}]{\includegraphics[width=.33\textwidth]{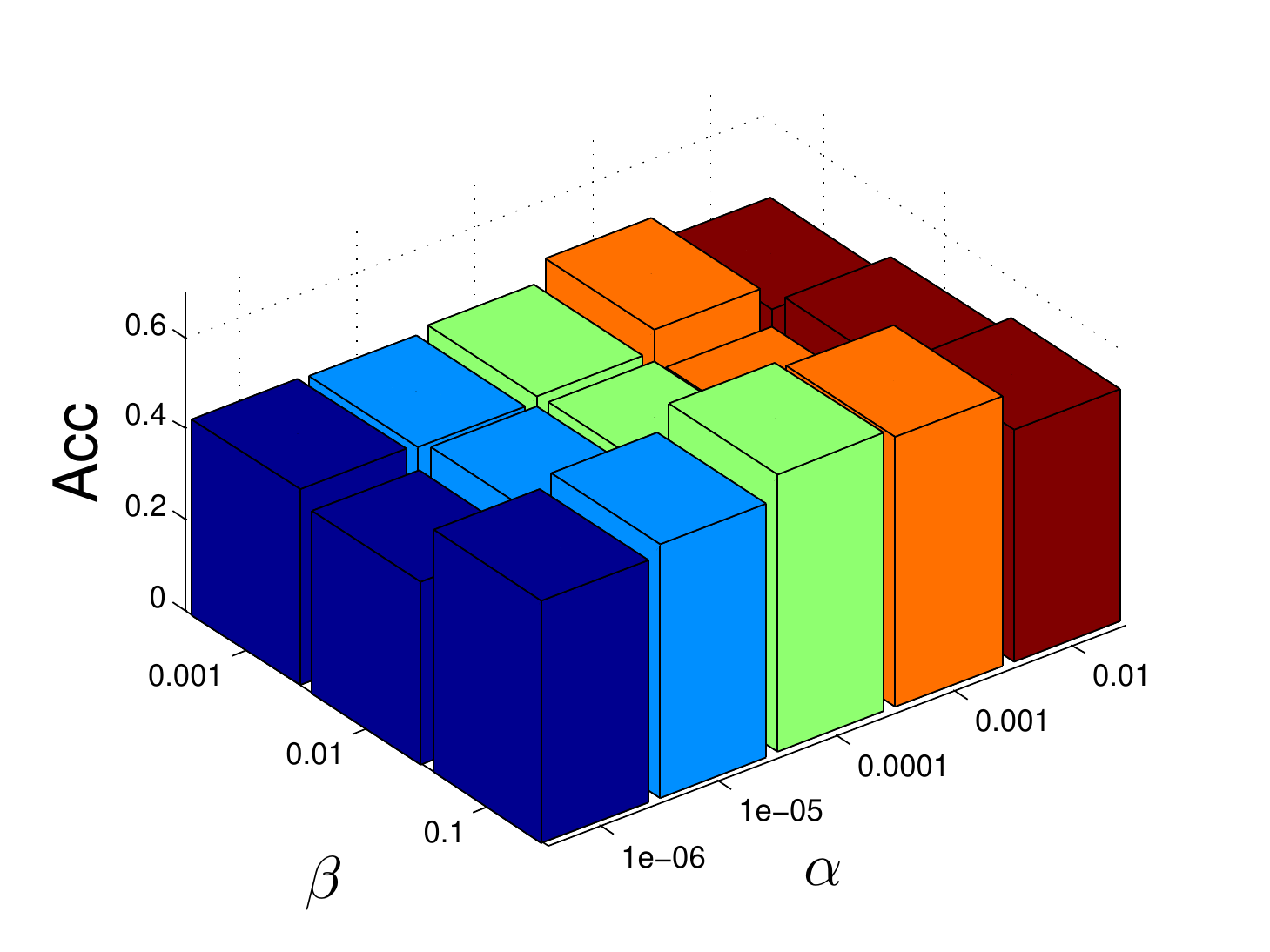}}
\caption{Parameter influence on accuracy for YALE data set.\label{yalepara}}
\end{figure*}
For single kernel methods, we run downloaded kernel k-means (KKM) \cite{scholkopf1998nonlinear}, spectral clustering (SC) \cite{ng2002spectral}, robust kernel k-means (RKKM) \cite{du2015robust}, and SCSK on each kernel separately. To demonstrate the advantage of our unified framework, we also implement three separate steps method (TSEP), i.e., learn the similarity matrix by (\ref{kernelsparse}), spectral clustering, k-means (repeat 20 times). And we report both the best and the average results over all these kernels. 

In addition, we also implement the recent simplex sparse representation (SSR) \cite{huang2015new} method and robust affinity graph construction methods by using random forest approach: ClustRF-u and ClustRF-a \cite{zhu2014constructing}. ClustRF-u assumes all tree nodes are uniformly important, while ClustRF-a assigns an adaptive weight to each node. Note that these three methods can only process data in the original feature space. Moreover, ClusteRF has a high demand for memory and cannot process high dimensional data directly. Thus we follow the authors' strategy and perform PCA on TR11, TR41, and TR45 to reduce the dimension. We use different numbers of dominant components and report the best clustering results. Nevertheless, we still cannot handle TDT2 data set with them.

For multiple kernel methods, we implement our proposed method and directly use the downloaded programs for the methods in comparison on a combination of these 12 kernels: 

MKKM\footnote{http://imp.iis.sinica.edu.tw/IVCLab/research/Sean/mkfc/code}. The MKKM \cite{huang2012multiple} extends k-means in a multiple kernel setting. However, it imposes a different constraint on the kernel weight distribution. 

AASC\footnote{http://imp.iis.sinica.edu.tw/IVCLab/research/Sean/aasc/code}. The AASC \cite{huang2012affinity} is an extension of spectral clustering to the situation when multiple affinities exist. It is different from our approach since our method tries to learn an optimal similarity graph.

RMKKM\footnote{\label{note1}https://github.com/csliangdu/RMKKM}. The RMKKM \cite{du2015robust} extends k-means to deal with noise and outliers in a multiple kernel setting.

SCMK. Our proposed method of spectral clustering with multiple kernels. For the purpose of reproducibility, the code is publicly available\footnote{https://github.com/sckangz/AAAI18}.

For our method, we only need to run once. For those methods that involve K-means, we follow the strategy suggested in \cite{yang2010image}; i.e., we repeat clustering 20 times and present the results with the best objective values. We set the number of clusters to the true number of classes for all clustering algorithms. 

\subsection{Results}
We present the clustering results of different methods on those benchmark data sets in Table \ref{clusterres}. In terms of accuracy, NMI and Purity, our proposed methods obtain superior results. The big difference between the best and average results confirms that the choice of kernels has a huge influence on the performance of single kernel methods. This motivates our extended model for multiple kernel learning. Besides, our extended model for multiple kernel clustering usually improves the results over our model for single kernel clustering. 

Although the best results of the three separate steps approach are sometimes close to our proposed unified method, their average values are often lower than our method. We notice that random forest based affinity graph method achieves good performance on image data sets. This observation can be explained by the fact that ClustRF is suitable to handle ambiguous and unreliable features caused by variation in illumination, face expression or pose on those data sets. On the other hand, it is not effective for text data sets. In most cases, ClustRF-a behaves better than ClustRF-u. This justifies the importance of considering neighbourhood-scale-adaptive weighting on the nodes.
\subsection{Parameter Sensitivity}

There are three parameters in our model: $\alpha$, $\beta$, and $\gamma$. We use YALE data set as an example to demonstrate the sensitivity of our model SCMK to parameters. As shown in Figure \ref{yalepara}, our model is quite insensitive to $\alpha$ and $\beta$, and $\gamma$ over wide ranges of values.  In terms of NMI and Purity, we have similar observations. 

\section{Conclusion}
\label{conclusion}
In this work, we address two problems existing in most classical spectral clustering algorithms, i.e., constructing similarity graph and relaxing discrete constraints to continuous one. To alleviate performance degradation, we propose a unified spectral clustering framework which automatically learns the similarity graph and discrete labels from the data. To cope with complex data, we develop our method in kernel space. A multiple kernel approach is proposed to solve kernel dependent issue. Extensive experiments on nine real data sets demonstrated the promising performance of our methods as compared to existing clustering approaches.
\section{Acknowledgments}
This paper was in part supported by Grants from the
Natural Science Foundation of China (No. 61572111), the
National High Technology Research and Development Program
of China (863 Program) (No. 2015AA015408),
a 985 Project of
UESTC (No.A1098531023601041) and a Fundamental
Research Fund for the Central Universities of China (No. A03017023701012). 
\bibliographystyle{named}
\bibliography{ref}

\end{document}